# Insect pest image detection and recognition based on bio-inspired methods

*Loris Nanni[a], Gianluca Maguolo[a,*], Fabio Pancino[a]*

[a]*University of Padova, via Gradenigo 6, Padova 35131, Italy*

[*]*Corresponding author*

**Emails**:

*loris.nanni@unipd.it*

*gianluca.maguolo@phd.unipd.it*

*fabio.pancino@studenti.unipd.it*

ARTICLE INFO

ABSTRACT

Insect pests recognition is necessary for crop protection in many areas of the world. In this paper we propose an automatic classifier based on the fusion between saliency methods and convolutional neural networks. Saliency methods are famous image processing algorithms that highlight the most relevant pixels of an image. In this paper, we use three different saliency methods as image preprocessing and create three different images for every saliency method. Hence, we create $3 \times 3 = 9$ new images for every original image to train different convolutional neural networks. We evaluate the performance of every preprocessing/network couple and we also evaluate the performance of their ensemble. We test our approach on both a small dataset and the large IP102 dataset. Our best ensembles reaches the state of the art accuracy on both the smaller dataset (92.43%) and the IP102 dataset (61.93%), approaching the performance of human experts on the smaller one. Besides, we share our MATLAB code at: https://github.com/LorisNanni/.

## 1. Introduction

Insect pests are one of the main causes of crop damages all over the world. The prevention of these damages could avoid the loss of a large portion of the harvest and allow a higher efficiency of the agriculture. Not only pests might reduce the harvest, they might also damage the machinery and the equipment (Kandalkar, Deorankar and Chatur, 2014). Reducing such a problem would lead to an economic growth of the whole sector and would require a lower amount of natural resources to produce the same amount of food. The first step to prevent crop damages due to insect pests is the ability to recognize and classify insects, in order to discriminate between the safe ones and the dangerous ones. Since this task requires a continuous and expensive monitoring, there has been a growing interest in automatic pests classification in recent years (Xia et al., 2018).

The development of image classification (Miranda, Gerardo and Tanguilig III, 2014)(Gondal and Khan, 2015) is offering new opportunities to researchers to develop increasingly automated systems capable of recognizing any type of object with a high degree of accuracy. In this paper we propose a new method to classify insect pests based on convolutional neural networks (CNNs) and saliency methods. Saliency methods are image processing algorithms that highlight the most relevant part of an image (Itti, Koch and Niebur, 1998). The idea behind these methods comes from the observation that the human eye does not focus on all its field of view, but accurately discriminates between its relevant and not relevant parts. Saliency methods try to mimic how the human eye selects the most relevant regions that it sees. Using these algorithms, we were able to create new labelled samples by discarding the pixels with low importance. In this way we augmented the image data and we used the new samples to train multiple networks to build an ensemble. As shown by (Nanni, Brahnam and Maguolo, 2019), creating an ensemble of different CNNs with the same architecture, trained on different augmented data, is an effective way to exploit data augmentation. In our experiments, we also create larger ensembles including different architectures. We show that these ensembles can classify pest images reaching a state of the art performance in two different datasets. To the best of our knowledge, we are the first to prove that using saliency methods for data augmentation leads to a large improvement in the performance.

The rest of the paper is organized as follows: in Section 2 we summarize some of the most relevant papers in the literature dealing with pests classification and with saliency methods. In Section 3 we introduce the datasets and the saliency methods that we use in this paper. Section 4 contains a brief introduction to CNNs and a description of the 5 architectures that we trained. In Section 5 we explain the details of our training protocol. The performance of our networks in Section 6, while, finally, some comments on these results are in Section 7.

## 2. Related work

### *2.1. Insects classification*

Insect pests classification has recently gained the attention of the academic community thanks to the great advances in computer vision. Several labelled datasets are available for classification, however most of them contain only a small number of samples, compared to the usual size required by deep neural networks. (Venugoban and Ramanan, 2014) proposed a very small dataset containing 200 samples divided in 10 classes. (Xie *et al.*, 2015) proposed a larger dataset made by 1440 samples evenly divided into 24 classes of insect pests that infest crop. (Deng *et al.*, 2018) proposed a novel dataset containing 563 images divided in 10 different classes. They trained a Support Vector Machine (SVM) on hand-crafted features to classify their dataset. Training support vector machines on hand-crafted features was the common approach, especially before the rise of deep learning and was still competitive on small datasets also after the advent of neural networks (Rani and Amsini, 2016). However, in recent years, CNNs outperformed every previous technique in tasks like image classification, image segmentation and object recognition, especially on large datasets. An attempt to automatically classify pests using CNNs was made by (Dawei *et al.*, 2019). The authors used a pretrained version of AlexNet (Krizhevsky, Sutskever and Hinton, 2012) to classify a portion of the dataset used in (Deng *et al.*, 2018) using transfer learning. In order to have a baseline comparison, they calculated the accuracy rate of six human experts on the same dataset. Their network managed to reach a higher accuracy than four of the six human experts. More recently, (Wu *et al.*, 2019) collected and shared the IP102 dataset, which contains over 75000 pest images divided in 102 classes, providing a useful benchmark dataset for researchers. Their effort on the collection and labelling of such a large number of images shows the importance of having a high-performing automatic pest classifier. (Ren, Liu and Wu, 2019) trained different convolutional neural networks on this dataset and proposed their own architecture for the classification.

*2.2. Saliency methods*

Saliency methods identify the most relevant regions of an image. They are currently used in many computer vision tasks to have unsupervised information about which pixels are important for a given task. (Wang *et al.*, 2016) used saliency methods in a 3D video recognition pipeline. (Cornia *et al.*, 2018) used saliency maps to create an attention mechanism for image captioning. In the field of classification (Han and Vasconcelos, 2010) used saliency methods for object recognition and classification, inspired by the idea of simulating a biological attention mechanism. (Murabito *et al.*, 2018) created their own saliency method explicitly designed for scene classification, since the peculiarity of the task require seeing the image as a whole and usual methods were not suitable. (Flores *et al.*, 2019) used saliency maps to highlight the most relevant pixels for fine-grained object recognition with few training data. This allowed their neural network to focus only on the relevant part of the image, which helped the, to overfitting on that task. In the next section, we shall describe more accurately the saliency methods that we shall use in this paper.

## 3. Materials and methods

### *3.1. Datasets*

We use the same dataset proposed in (Deng *et al.*, 2018). It is made of ten different classes of pests that are mainly found in tea plants and in other plants scattered between Europe and Central Asia. The division of the dataset can be found in Table 1. This collection of images comes from various online sources, one of which is Mendeley Data, which contains photos taken with a Single Lens Reflex (SLR). The other images were taken from Insert Images, IPM images, Dave's Garden and other sources. Some image samples can be found in Figure 1.

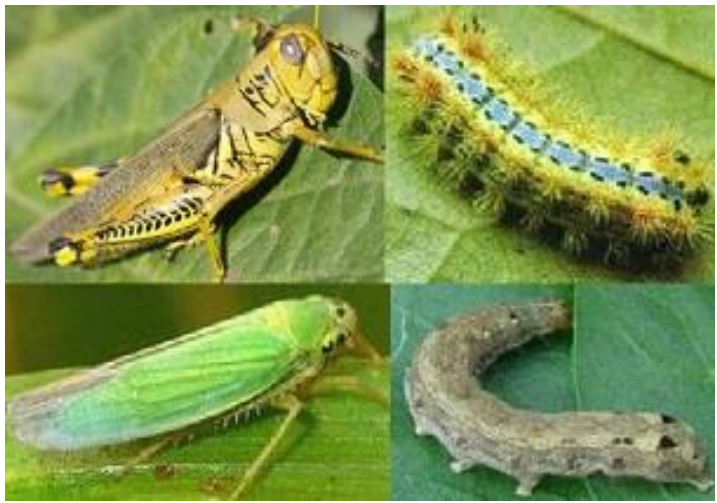

Figure 1. Image samples from Deng et al.

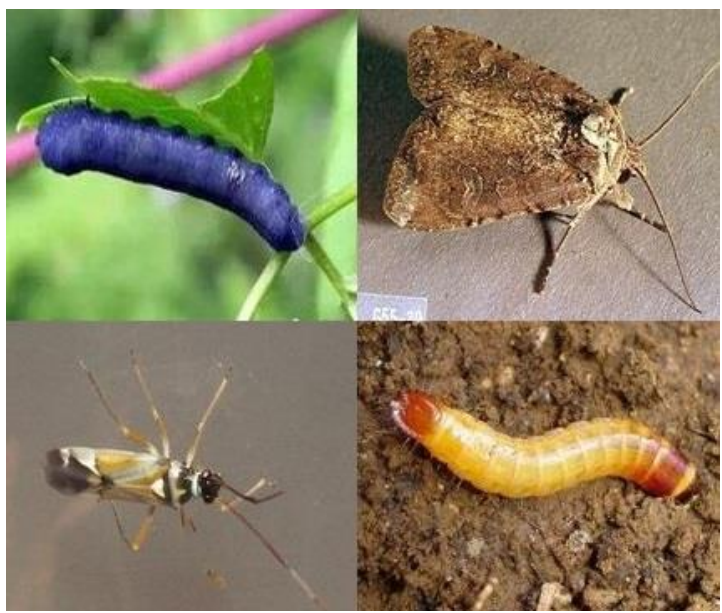

Figure 2. Image samples from IP102

Table 1 – Composition of the dataset

| Species | Number of Samples |
|---|---|
| Locusta migratoria | 72 |
| Parasa lepida | 59 |
| Gypsy moth larva | 40 |
| Empoasca flavescens | 41 |
| Spodoptera exigua | 68 |
| Chrysocus chinensis | 50 |
| Laspeyresia pomonella larva | 50 |
| Spodoptera exigua larva | 56 |
| Atractomorpha sinensis | 62 |
| Laspeyresia pomonella | 65 |

We also tested our approach on the IP102 dataset, a much larger dataset containing 75222 images divided in 102 different classes. The dataset is split in 45095 images for training, 7508 for validation and 22169 for testing. The least represented class in the dataset only has 71 samples, while the most represented class contains more than 5740 samples, hence the dataset is highly unbalanced. It also worth mentioning that, in this dataset, insects of very different age are included in every class, which is a challenging problem since the shape of an insect, like a butterfly, might change completely during his life. Some examples are shown in Figure 2.

### 3.2. Saliency Methods

Saliency maps (Itti, Koch and Niebur, 1998) are functions that take an image as an input and output those locations where the image has its most relevant features. It is usually made of three different steps: a linear feature extraction, the application of a non linearity (activation) to the extracted features, and, at last, the fusion of the features. In our paper, we extract several saliency maps from the images in our dataset using three different algorithms, namely, Graph-Based Visual Saliency (GBVS), Cluster-based Saliency Detection (COS) and Spectral Residual (SPE). In Figures 3-5 one can see which pixels of the original images are highlighted by the saliency methods. Every saliency method can be used to create a binary mask by setting to 1 all the pixels whose value is above a threshold and setting to 0 all the others. For every image in the dataset, we use a saliency method to create three different images: foreground (FG), region of interest (ROI) and foreground region of interest (FG-ROI). The FG image consists in the multiplication of the original image and its binary mask. This operation sets the foreground of the image to black. The FG-ROI image is a smaller image that only contains the insect. It discards the columns and the rows of the image where the binary mask is equal to zero almost everywhere. In the ROI image we both turn to black the foreground and extract a region of interest.

*3.2.1. GBVS*

Graph-Based Visual Saliency (Simonyan, Vedaldi and Zisserman, 2013) is an algorithm to combine different feature maps. The idea is to use the feature maps to create a dissimilarity measure on the pixels of the image. This measure is used to create a Markov chain on the fully-connected graph made by the pixels of the image. The transition probability $p\big((i,j),(k,l)\big)$ from $(i,j)$ to $(k,l)$ on this graph is proportional to the multiplication of the dissimilarity of the two pixels and a decreasing function of their distance:

$$p\big((i,j),(k,l)\big) = d\big((i,j)\,||\,(k,l)\big) \cdot F(i-k, j-l) \tag{1}$$

where

$$d\big((i,j)\,||\,(k,l)\big) = \left|\log\left(\frac{M(i,j)}{M(k,l)}\right)\right| \tag{2}$$

and

$$F(a,b) = \exp\left(-\frac{a^2 + b^2}{2\sigma^2}\right) \tag{3}$$

and $M$ is a feature map. The transition probability must also be normalized in order to sum to 1. The saliency map is given by the stationary probability distribution associated to the Markov chain. The most relevant pixels found by this method are cluster of pixels that have a large dissimilarity. To give an example of how the algorithm works, let $I$ be a small image with 9 pixels and suppose that $M(i,j)$ is constant for every pixel except $M(1,1)$. Then, the transition probability towards $(1,1)$ will be very large, hence the stationary distribution will have a large value in $(1,1)$. Hence this point will be highlighted by the saliency map. This toy example shows how this method highlights small groups of pixels which are very different from the ones that are in the same region.

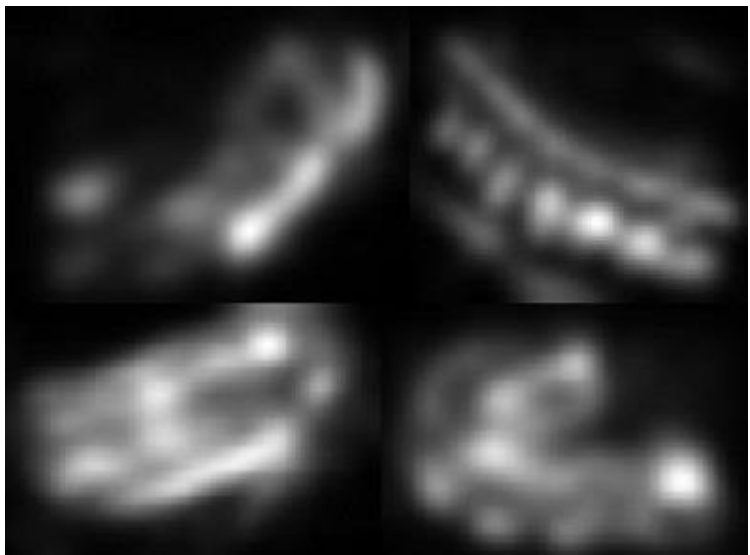

Figure 3. Binary masks for GBVS.

*3.2.2. Cluster-based Saliency Detection (COS)*

The co-saliency map can be used in various vision applications. It is used to discover the common saliency on multiple images. Co-saliency detection uses the repetitiveness property as additional constraint, and discovers the common salient object in the images.

The idea of this method is presented in (Fu, Cao and Tu, 2013). Given a set of images, the authors use a two layer clustering; initially, one layer groups the pixels on each image (single image), and the other layer associates the pixels on all the images (multi-image). Then the saliency cues (Contrast cue, Spatial cue, Corresponding cue) are computed for each cluster, and the cluster-level saliency is measured. The measured features include

- Contrast cue: the advantage of the contrast cue is the ability of identifying the rarest pixel clusters within the image. It is very efficient in cases where there is a single subject. It loses efficiency with the increase of similar subjects within the image.
- Distance from the image center (on single/multi-image). Spatial cue limits the low performance of the contrast cue when more subjects appear in the background. This indicator is based on the spatial distribution of the pixels clusters and is able to highlight the most salient cluster of the image through the calculation of the Euclidean distance between the centroid of the cluster and the center of the image.
- the repetitiveness is computed by measuring how the distribution of clusters varies, giving a higher score to those that appear more frequently, in order to be able to identify with better approximation the most salient clusters.

At last, based on these cluster-level cues, our method computes the saliency value for each pixel, which is used to generate the final saliency map.

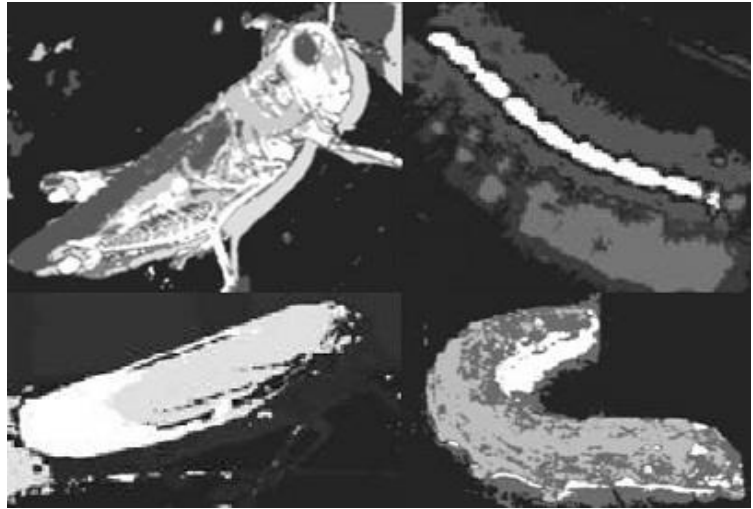

Figure 4. Binary masks for COS.

*3.2.3. Spectral residual (SPE)*

Spectral residual is a method that allows us to derive the saliency map of an input image through the analysis of the residual spectrum (Hou and Zhang, 2007).

The Fourier Transform decomposes an image into its sine and cosine components. The output of the transformation represents the image in the Fourier or frequency domain, while the input image is the spatial domain equivalent. In the Fourier domain image, each point represents a particular frequency contained in the spatial domain image.

The statistical average of a signal, defined as its frequency content, is called its spectrum. The scale invariance property states that the amplitude $A(f)$ of the averaged Fourier spectrum, where $f$ is a sinusoidal component of frequency, is a feature that does not change in the object if the scale changes and it obeys a distribution:

$$E\{A(f)\} \propto \frac{1}{f} \qquad (4)$$

The log spectrum representation $L(f)$ of an image is defined by $L(f) = \log(A(f))$. The log spectrum allows to see the low frequencies of the Fourier scale. Besides, the log spectra of different images share similar trends, though each containing statistical singularities.

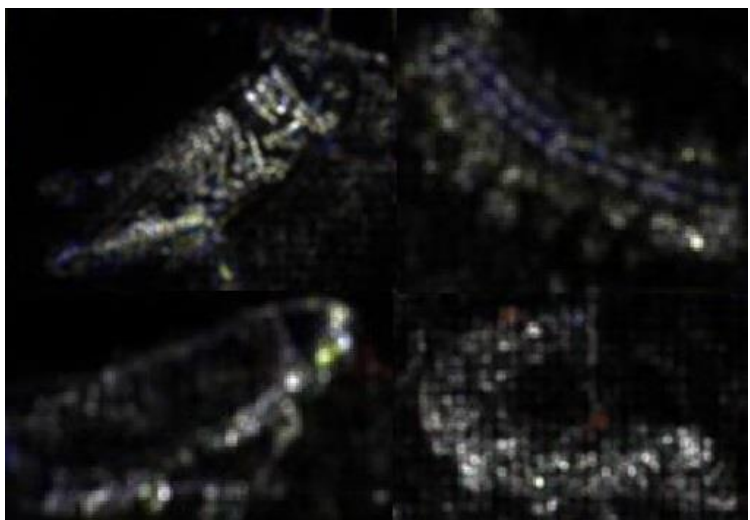
Figure 5. Binary masks for SPE.

The content of the residual spectrum can be interpreted as the unexpected portion of the image. SPE consists in highlighting those singularities.

## 4. Convolutional Neural Networks

Convolutional Neural Networks (CNNs) (Krizhevsky, Sutskever and Hinton, 2012) are a class of multi-layer neural networks, designed to recognize visual patterns directly from pixel images with minimal pre-processing.

The architecture of such networks can be summarized in a feature extractor and a classifier that are trained end-to-end. The feature extractor is made by many convolutional and pooling layers. Convolutional layers performs weighted convolutions between their inputs and their learnable weights. Thus, they find local patterns in the input. Pooling layers are non-trainable layers that reduce the dimensionality of their input by locally mapping a little square in the input into a single number. The classifier is usually made by one or more fully connected layers and a softmax function. However, light networks designed for mobile applications do not use fully connected layers because they contain the largest number of parameters of the network.

The CNNs used in this paper are: AlexNet, GoogLeNet, ShuffleNet, MobileNetv2, DenseNet201.

*4.1.1. AlexNet (AN)*

AlexNet (Krizhevsky, Sutskever and Hinton, 2012) was designed in 2012 and it managed to outperform every previous algorithm in ImageNet classification. It contains eight learnable layers: the first five are convolutional and the remaining three are fully-connected. Every learnable layer is followed by a Rectified Linear Unit (ReLU) as activation function. Among the convolutional layers, max-pooling layers are used to reduce the dimensionality of the hidden layers. The output of the last fully-connected layer is fed into a softmax function that produces a probability distribution over the possible output classes.

*4.1.2. GoogLeNet (GL)*

GoogleNet (Szegedy *et al.*, 2015) architecture consists of a 22 layer CNN. It only has 4 million parameters, which is a very low number compared with the 60 million parameters of AlexNet. The main feature of GoogLeNet is the use of inception layers. They are made of different convolutional layers with different size, whose outputs are combined at the end of the module. The idea is that filters of different size can spot different patterns of the input.

*4.1.3. ShuffleNet (SN)*

ShuffleNet (Zhang *et al.*, 2018) is a very light network that uses grouped convolutions and channel shuffle to have a very low complexity. Grouped convolution are $1 \times 1$ convolutions that only consider a subset of the channels of the hidden layer. In this way the number of multiplications is strongly reduced. Then, in order to increase the communication between the neurons, the channels of the output are shuffled. The result is that the network is 13 times faster than AlexNet, maintaining similar accuracy.

*4.1.4. MobileNetv2 (MN)*

MobileNetv2 (Sandler *et al.*, 2018) is a light convolutional network for mobile applications. It contains several depthwise separable convolutions, which can be thought as convolutional layers where the 3D weights tensor is factorized in one 2D tensor and a 1D tensor, requiring much less memory. This kind of layers is often used in light convolutional networks. With respect to other networks, MobileNetv2 has fewer nonlinearities. The authors give an interpretation on why this should give better results in their network and state that the same network performed worse when they tried to add more linearities. Besides, this network has inverted skip connections. This means that the hidden layers connected through skip connections are low dimensional, reducing the number of operations made by the network.

*4.1.5. DenseNet201 (DN)*

DenseNet201 (Huang *et al.*, 2017) is large and high-performing convolutional network. Its name comes from the fact that every layer is connected to all the previous layers. This architecture helps gradient flow and encourages feature reuse. DenseNet201 is competitive with other state-of-the-art networks on ImageNet while having less parameters than AlexNet.

**5. Training**

We trained every CNN using Stochastic Gradient Descent with Momentum. We fine-tuned the pre-trained networks introduced in Section 4 by replacing their last fully connected layer with a smaller one whose output size is the number of classes of the training dataset. In order to learn faster in the new layers than in the transferred ones, we increased the relative learning rate of the last fully connected layer by a factor of 20. In the inner layers we used a learning rate of 0.0001.

In both datasets we used standard data augmentation techniques: we randomly reflected the images along both axis, rotated them by an angle between [-10,10], and translated them along both axes, by a number of pixels between [0,5]. Finally we randomly scale the images in both axis by a value that varies between [1,2]. In the smaller dataset we used mini-batches of size 30 and trained the networks for 30 epochs. In IP102 we used mini-batches of size 128 and trained the network for at most 60 epochs, but using the validation set for early stopping.

The training set of the small dataset consists in 20 images randomly extracted from every class, while the remaining ones are in the test set. This protocol and the split are the same used in (Deng *et al.*, 2018). This splitting is performed five times, then the average results are reported. For the IP102 dataset we used the same training, validation and test split proposed by their creators.

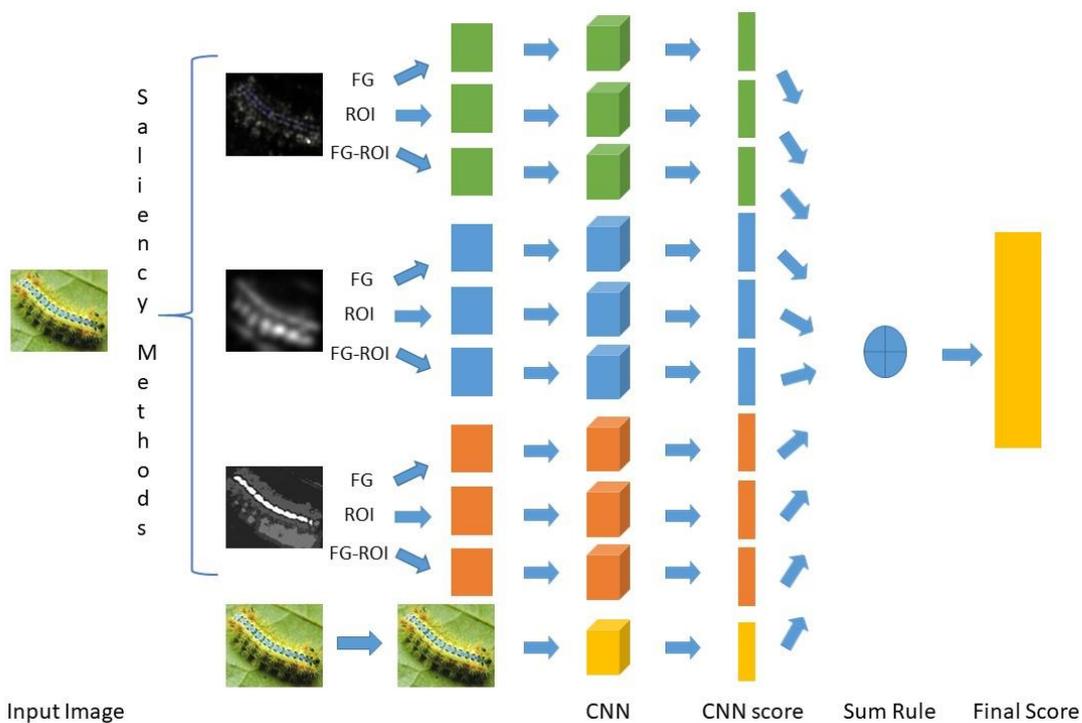

Figure 6. Scheme of the Method

For every original dataset, we created nine more different datasets. For every saliency method we cretaed: a Foreground (FG) dataset where the background of every insect is turned to black; a Foreground Region Of Interest (FG-ROI) dataset made by the portions of the original images that contain the insects; a Region Of Interest (ROI) dataset the both turns to black the foreground and extracts a region of interest.

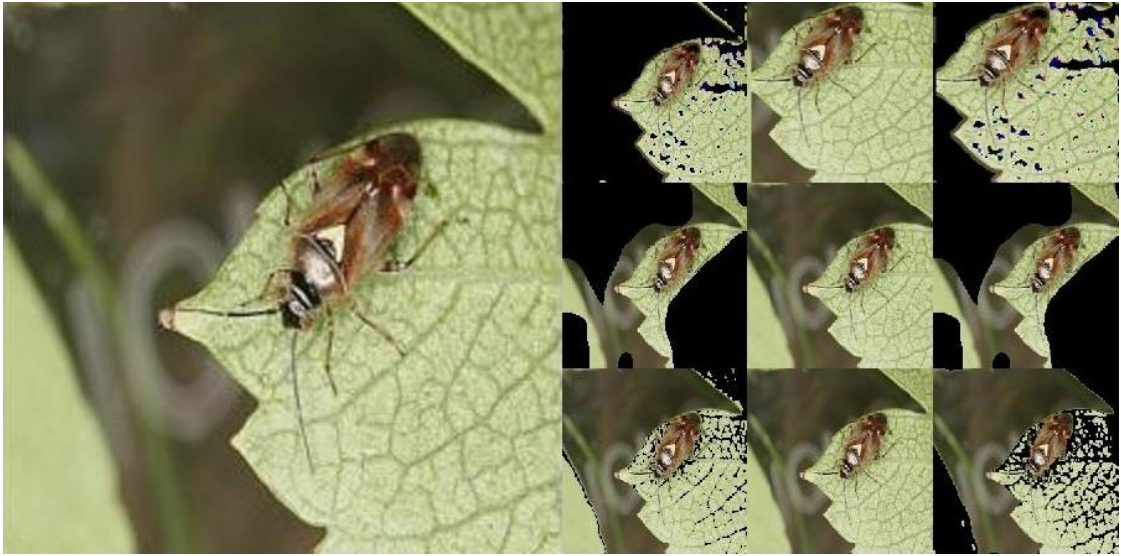

Figure 7 – Comparisons between the original image and the extracted images.

We provide some example images in Figure 7. The images on the left is the original one. On the right, we show the 9 extracted images. In the three rows, one can see respectively SPE, GBVS and COS. On the three columns one can see FG, FG-ROI and ROI. For both the small dataset and IP102, we trained every network on 10 different datasets: the original one and the other 9 obtained using the saliency maps of Section 3 as image preprocessing. We summarize our method in Figure 6. We report the performance of every protocol and we also evaluate the performance of the ensemble of these networks using the sum rule on the softmax output.

## 6. Experimental Results

The results reported in Tables 2-4 include the training of the single CNNs on the different datasets. The row 'OriginalImage' shows the results obtained by the CNNs on the original dataset. The rows named after saliency methods contain the performance of the networks after the saliency method has been applied. 'FusionSum' is the performance of the ensemble created by the same network trained on all the datasets, while 'AllSum' is the ensemble created with all the networks trained.

The other ensembles reported in Tables 2-4 are:

- The row 'AllSum\Spectral' contains the performance of the ensemble of all the networks but the ones trained on the SPE datasets, which are the worst performing. This shows that even those networks are useful when included in the ensemble, probably because the information brought by those networks is somehow independent from the information brought by the others.

- The row 'FusionSum\FG_ROI contains the performance of the ensemble FusionSum but the ones trained on the FG_ROI images.
- The row 'AllSum\DN contains the performance of the ensemble of all the networks but the ones based on densenet201, it is an ensemble of light CNN.
- The row 'AllSum\FG_ROI\DN contains the performance of the ensemble of all the networks but the ones based on densenet201 or FG_ROI.
- The row 'AllSum\Spectral\DN contains the performance of the ensemble of all the networks but the ones based on densenet201 or on the SPE datasets.

We also report the results of other authors on the same datasets.

Table 2 – Accuracy Rates on the (Deng *et al.*, 2018) dataset

| SMALL | | AN | GN | MN | SN | DN |
|---|---|---|---|---|---|---|
| COS | FG | 78.67 | 79.78 | 76.19 | 77.68 | 85.03 |
| | ROI | 77.35 | 79.12 | 74.97 | 76.80 | 83.92 |
| | FG_ROI | 83.15 | 87.46 | 84.14 | 86.46 | 88.34 |
| GBVS | FG | 79.61 | 82.98 | 76.63 | 80.55 | 83.04 |
| | ROI | 77.57 | 81.16 | 76.80 | 78.56 | 77.62 |
| | FG_ROI | 81.16 | 84.14 | 81.44 | 83.59 | 86.74 |
| SPE | FG | 73.26 | 76.63 | 65.97 | 69.78 | 69.45 |
| | ROI | 73.31 | 74.42 | 64.97 | 69.34 | 69.06 |
| | FG_ROI | 86.69 | 87.73 | 83.31 | 84.81 | 88.51 |
| OriginalImage | | 83.76 | 85.69 | 81.55 | 86.52 | 90.22 |
| FusionSum | | 88.23 | 90.77 | 89.72 | 91.66 | 92.10 |
| FusionSum\FG_ROI | | 86.30 | 87.68 | 85.97 | 89.06 | 91.60 |
| AllSum\DN | | 91.88 | | | | |
| AllSum\FG_ROI\DN | | 90.83 | | | | |
| AllSum\Spectral\DN | | 89.89 | | | | |
| AllSum | | **92.43** | | | | |
| (Deng *et al.*, 2018) | | 85.50 | | | | |

Table 3: Human experts performance on the small dataset

| Expert | 1 | 2 | 3 | 4 | 5 | 6 |
|---|---|---|---|---|---|---|
| Performance | 0.96 | 0.96 | 0.92 | 0.91 | 0.90 | 0.82 |

Table 4 – Accuracy Rates on the IP102 dataset

| LARGE | | AN | GN | MN | SN | DN |
|---|---|---|---|---|---|---|
| COS | FG | 44.08 | 47.31 | 48.28 | 45.47 | 52.45 |
| | ROI | 41.58 | 44.52 | 46.77 | 42.99 | 51.98 |
| | FG_ROI | 47.01 | 49.08 | 51.51 | 49.50 | 56.14 |
| GBVS | FG | 42.86 | 48.13 | 49.92 | 46.35 | 54.56 |
| | ROI | 45.77 | 49.54 | 50.79 | 48.08 | 55.65 |
| | FG_ROI | 49.75 | 51.97 | 53.82 | 51.27 | 57.86 |
| SPE | FG | 35.90 | 41.68 | 43.68 | 39.22 | 47.77 |
| | ROI | 37.73 | 41.00 | 43.17 | 39.77 | 47.26 |
| | FG_ROI | 48.91 | 51.01 | 52.16 | 50.07 | 56.29 |
| OriginalImage | | 51.79 | 53.80 | 55.35 | 52.45 | 58.76 |
| FusionSum | | 54.60 | 56.13 | 58.92 | 56.43 | **61.93** |
| FusionSum\FG_ROI | | 53.63 | 55.49 | 57.94 | 55.63 | 61.21 |
| AllSum\DN | | 59.65 | | | | |
| AllSum\Spectral\DN | | 59.73 | | | | |
| AllSum | | 61.44 | | | | |
| (Wu *et al.*, 2019) | | 49.50 | | | | |
| (Ren, Liu and Wu, 2019) | | 55.24 | | | | |

We also evaluated two more metrics, F-score and G-mean, as a comparison on the IP102 dataset, since it is highly unbalanced. We reported them in Tables 5. We reach the state of the art on the IP102 dataset also in these two metrics. The two metrics are calculated as the weighted average of the values of the one vs. all binary classifiers, where the weight is the fraction of samples of a class in the test set.

Table 5 – F-score and G-mean on IP102

|  | F-score | G-mean |
|---|---|---|
| FusionSum Densenet | **0.592** | **0.755** |
| (Wu *et al.*, 2019) | 0.401 | 0.315 |
| (Ren, Liu and Wu, 2019) | 0.541 | _____ |

# 7. Discussion

Among the preprocessed datasets, GBVS is the best one and COS comes right after. As expected, SPE is the worst performing method. The motivation of this result can be found in Figures 2-4 and in Figure 7, where one can see that SPE highlights very few pixels and does not consider important parts of the foreground. We guess that in those situations where the insect is very small this method often fails. In general, the saliency methods reduce the performance of the networks with respect of the original image. This could be because of a correlation between the insect and its foreground, which is cut by the saliency method. However, the ensemble of a specific network (FusionSUM) outperforms each single network.

As a comparison with the previous literature, in (Deng *et al.*, 2018), the authors train a SVM classifier and reach an accuracy of 85.5% on the small dataset, which is clearly outperformed by our best ensemble. To the best of our knowledge, the only other paper where the same dataset is used is (Dawei *et al.*, 2019). However, they only use a partial and unbalanced version of the dataset because they did not manage to collect all the original images. Besides, they perform a random train-test split where the 70% of the images were used for a single training. They reach a 93.84% using transfer learning with AlexNet, however the comparison is not fair. The same technique tested with our protocol yields a 83.76% accuracy. It is worth noticing that they also report the performance of human experts on their test set, which are reported in Table 3. Although the comparison is not fair, since the test sets are different, we can see that our results are nearly as good as the ones obtained by the human experts.

Our ensembles also reach the state of the art on the IP102 dataset. As a comparison, we report the results obtained by (Wu *et al.*, 2019), that trained some baseline networks by freezing all the layers except the last one, and by (Ren, Liu and Wu, 2019), that proposed their own architecture in their paper.

As we expected, our experiments show that DenseNet is the best architecture. It is much heavier than ShuffleNet and MobileNet and more recent than GoogleNet and AlexNet. The best ensemble on IP102 is the one containing all the DenseNets, since the performance of the other networks are not good enough to increase the performance of the DenseNets. Surprisingly, this was no longer true in the small dataset, where the best ensemble was made by taking all the networks. Maybe its small size and the absence of a validation set size led DenseNet to overfit. Apart from the DenseNets, MobileNet is the best network on IP102 and the worst one on the small dataset, among the networks trained on the original dataset. We also want to stress the fact that FUSION SUM improves every stand-alone network.

In Figure 8 we show some of the wrong classifications. In the first row there are the misclassified images. In the second and the third rows, we respectively show an image belonging to the predicted class and a different image belonging to the true class. In particular, the samples in the second and the third columns, the first and the third images belong to the same class, although they are very different. Also in the fourth column, the network classifies the first and the second images as they belonged to the same class. However, they are not the same species. This figure shows that the intra-class variation is very large and that a correct classification is very hard in these cases.

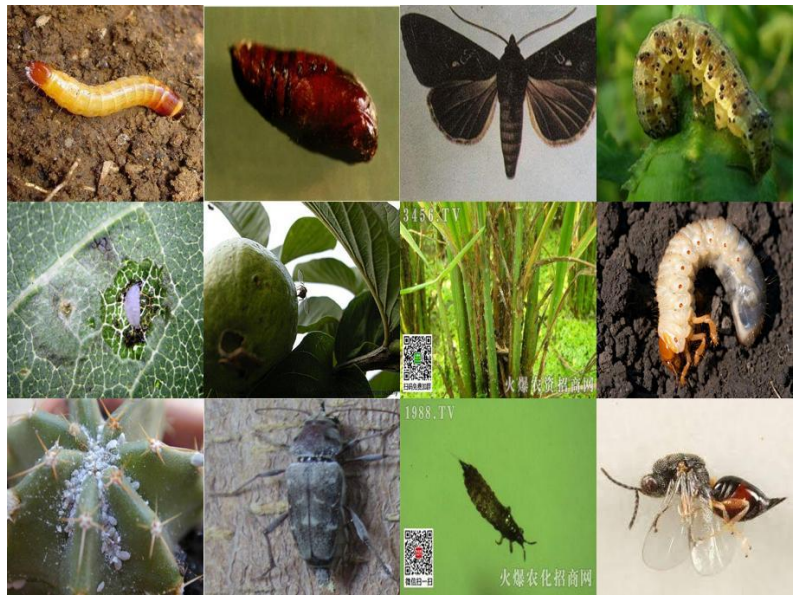

Figure 8 – Misclassifications and comparisons on the IP102 dataset.

## 8. Conclusions

The purpose of this paper was to create an automatic classifier of pest species. We explored the possibility to combine CNNs and saliency methods to create an ensemble of classifiers combined by the sum rule. We showed that training different CNNs on augmented data is a valuable tool for applying data augmentation. Besides, we showed that saliency methods can be used for augmenting the data. In particular, we trained 5 different CNNs architectures on the original dataset and on 9 synthetic datasets created using saliency methods as data augmentation. In this way, we had 50 CNNs trained on the same dataset. We evaluated their performance as stand-alone networks and we also evaluated several ensembles obtained using a subset of those networks. We tested our approach on two datasets that were very different in size and our best ensemble managed to establish a new state of the art on both of them, using three different evaluation metrics. Only (Dawei *et al.*, 2019) achieve better performances than we do, but with a simpler testing protocol. Their method trained and tested with our testing protocol performs worse than ours. Besides, in order to encourage the replication and the improvement of our methods, we share all our MATLAB code at https://github.com/LorisNanni/.

As future work, we shall investigate if different saliency methods can lead to a further improvement of the results. Besides, we are interested in applying active learning to this problem by downloading a large number of images from the web to select a small number of new images to be labelled that might sensibly improve the classification performance.

**Acknowledgement**

The authors wish to acknowledge the support of NVIDIA Corporation with their donation of the GPU used in this research.

**References**

Cornia, M. *et al.* (2018) 'Paying more attention to saliency: Image captioning with saliency and context attention', *ACM Transactions on Multimedia Computing, Communications, and Applications (TOMM)*. ACM New York, NY, USA, 14(2), pp. 1–21.

Dawei, W. *et al.* (2019) 'Recognition pest by image-based transfer learning', *Journal of the Science of Food and Agriculture*. Wiley Online Library, 99(10), pp. 4524–4531.

Deng, L. *et al.* (2018) 'Research on insect pest image detection and recognition based on bio-inspired methods', *Biosystems Engineering*. Elsevier, 169, pp. 139–148.

Flores, C. F. *et al.* (2019) 'Saliency for fine-grained object recognition in domains with scarce training data', *Pattern Recognition*. Elsevier, 94, pp. 62–73.

Fu, H., Cao, X. and Tu, Z. (2013) 'Cluster-based co-saliency detection', *IEEE Transactions on Image Processing*. IEEE, 22(10), pp. 3766–3778.

Gondal, M. D. and Khan, Y. N. (2015) 'Early Pest Detection from Crop using Image Processing and Computational Intelligence', *FAST-NU Research Journal ISSN*, pp. 2313–7045.

Han, S. and Vasconcelos, N. (2010) 'Biologically plausible saliency mechanisms improve feedforward object recognition', *Vision research*. Elsevier, 50(22), pp. 2295–2307.

Hou, X. and Zhang, L. (2007) 'Saliency detection: A spectral residual approach', in *2007 IEEE Conference on Computer Vision and Pattern Recognition*, pp. 1–8.

Huang, G. *et al.* (2017) 'Densely connected convolutional networks', in *Proceedings of the IEEE conference on computer vision and pattern recognition*, pp. 4700–4708.

Itti, L., Koch, C. and Niebur, E. (1998) 'A model of saliency-based visual attention for rapid scene analysis', *IEEE Transactions on Pattern Analysis & Machine Intelligence*. IEEE, (11), pp. 1254–1259.

Kandalkar, G., Deorankar, A. V and Chatur, P. N. (2014) 'Classification of agricultural pests using dwt and back propagation neural networks', *International Journal of Computer Science and Information Technologies*. Citeseer, 5(3), pp. 4034–4037.

Krizhevsky, A., Sutskever, I. and Hinton, G. E. (2012) 'ImageNet Classification with Deep Convolutional Neural Networks', *Advances in Neural Information Processing Systems 25 (NIPS 2012)*. doi: 10.1061/(asce)gt.1943-5606.0001284.

Miranda, J. L., Gerardo, B. D. and Tanguilig III, B. T. (2014) 'Pest detection and extraction using image processing techniques', *International Journal of Computer and Communication Engineering*. IACSIT Press, 3(3), p. 189.

Murabito, F. *et al.* (2018) 'Top-down saliency detection driven by visual classification', *Computer Vision and Image Understanding*. Elsevier, 172, pp. 67–76.

Nanni, L., Brahnam, S. and Maguolo, G. (2019) 'Data Augmentation for Building an Ensemble of Convolutional Neural Networks', in *Innovation in Medicine and Healthcare Systems, and Multimedia*. Springer, pp. 61–69.

Rani, R. U. and Amsini, P. (2016) 'Pest identification in leaf images using SVM classifier', *International Journal of Computational Intelligence and Informatics*, 6(1), pp. 248–260.

Ren, F., Liu, W. and Wu, G. (2019) 'Feature Reuse Residual Networks for Insect Pest Recognition', *IEEE Access*. IEEE, 7, pp. 122758–122768.

Sandler, M. *et al.* (2018) 'Mobilenetv2: Inverted residuals and linear bottlenecks', in *Proceedings of the IEEE Conference on Computer Vision and Pattern Recognition*, pp. 4510–4520.

Simonyan, K., Vedaldi, A. and Zisserman, A. (2013) 'Deep inside convolutional networks: Visualising image classification models and saliency maps', *arXiv preprint arXiv:1312.6034*.

Szegedy, C. *et al.* (2015) 'Going deeper with convolutions', in *Proceedings of the IEEE conference on computer vision and pattern recognition*, pp. 1–9.

Venugoban, K. and Ramanan, A. (2014) 'Image classification of paddy field insect pests using gradient-based features', *International Journal of Machine Learning and Computing*. IACSIT Press, 4(1), p. 1.

Wang, X. *et al.* (2016) 'Beyond frame-level CNN: saliency-aware 3-D CNN with LSTM for video action recognition', *IEEE Signal Processing Letters*. IEEE, 24(4), pp. 510–514.

Wu, X. *et al.* (2019) 'IP102: A Large-Scale Benchmark Dataset for Insect Pest Recognition', in *Proceedings of the IEEE Conference on Computer Vision and Pattern Recognition*, pp. 8787–8796.

Xia, D. *et al.* (2018) 'Insect detection and classification based on an improved convolutional neural network', *Sensors*. Multidisciplinary Digital Publishing Institute, 18(12), p. 4169.

Xie, C. *et al.* (2015) 'Automatic classification for field crop insects via multiple-task sparse representation and multiple-kernel learning', *Computers and Electronics in Agriculture*. Elsevier, 119, pp. 123–132.

Zhang, X. *et al.* (2018) 'Shufflenet: An extremely efficient convolutional neural network for mobile devices', in *Proceedings of the IEEE Conference on Computer Vision and Pattern Recognition*, pp. 6848–6856.